\documentclass[conference]{IEEEtran}
\usepackage{times}
\usepackage{epsfig}
\usepackage{psfrag}
\usepackage{graphicx}
\usepackage{amsmath}
\usepackage{amssymb}
\usepackage{mathrsfs}
\usepackage{enumerate}
\usepackage[lined,boxed,commentsnumbered]{algorithm2e}
\usepackage{multirow}

\usepackage[lined,boxed,commentsnumbered]{algorithm2e}

\begin{document}

\title{Layered Interpretation of Street View Images}

\author{\authorblockN{Ming-Yu Liu\authorrefmark{1},
Shuoxin Lin\authorrefmark{2},
Srikumar Ramalingam\authorrefmark{1}, and
Oncel Tuzel\authorrefmark{1}
\authorblockA{\authorrefmark{1}Mitsubishi Electric Research Labs (MERL), Cambridge, Massachusetts, USA\\
Email: \{mliu,ramalingam,oncel\}@merl.com}
\authorblockA{\authorrefmark{2}University of Maryland College Park, Maryland, USA\\
Email: slin07@umd.edu}}
}

\maketitle

\begin{abstract}
We propose a layered street view model to encode both depth and semantic information on street view images for autonomous driving. Recently, stixels, stix-mantics, and tiered scene labeling methods have been proposed to model street view images. We propose a 4-layer street view model, a compact representation over the recently proposed stix-mantics model. Our layers encode semantic classes like ground, pedestrians, vehicles, buildings, and sky in addition to the depths. The only input to our algorithm is a pair of stereo images. We use a deep neural network to extract the appearance features for semantic classes. We use a simple and an efficient inference algorithm to jointly estimate both semantic classes and layered depth values. Our method outperforms other competing approaches in Daimler urban scene segmentation dataset. Our algorithm is massively parallelizable, allowing a GPU implementation with a processing speed about 9 fps.
\end{abstract}

\IEEEpeerreviewmaketitle

\section{Introduction}

\begin{figure}[!t]
\centering
\includegraphics[width=1.0\columnwidth]{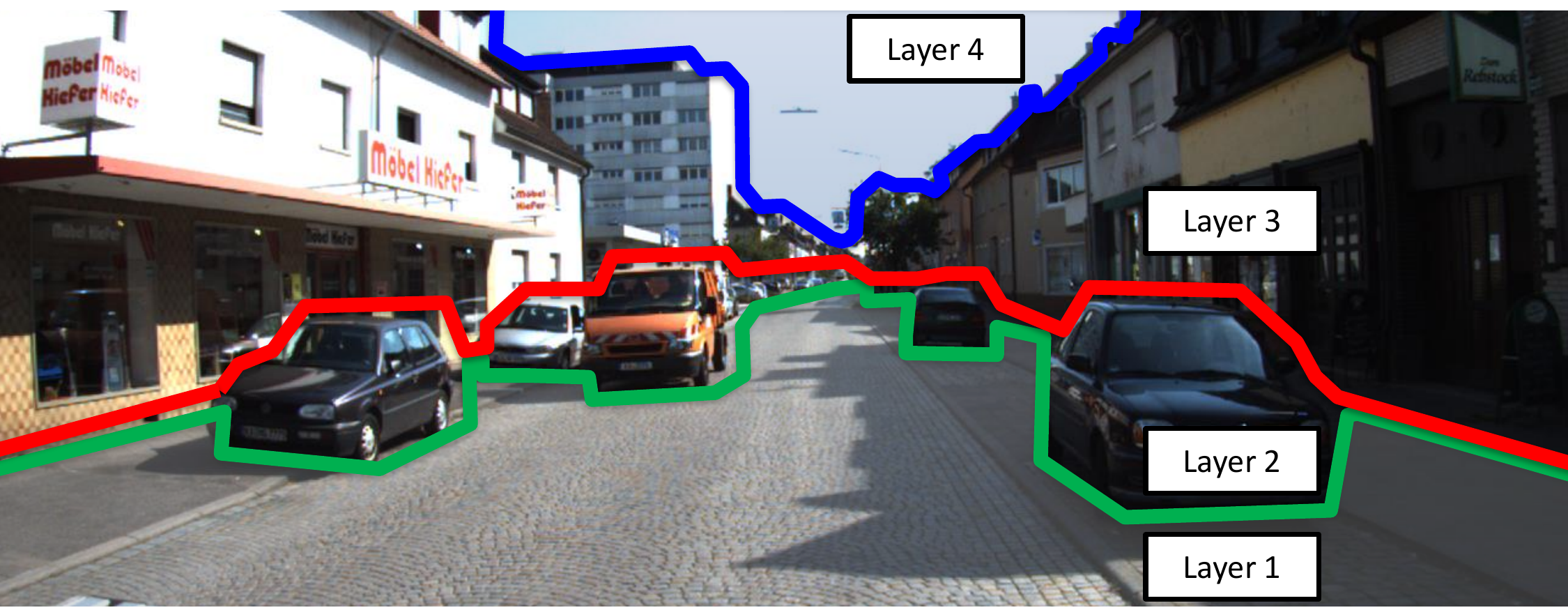}
\caption{We propose a layered interpretation algorithm for street scene understanding. We use four layers to represent every street scene image. We encode the semantic classes of several important object classes in the region between these layers. These classes include ground, pedestrians, cars, building, and sky. In addition to semantic classes, we also model and compute the layer-aware depth of the scene. In other words, the depth increases for pixels as we move to the top of the image along an image column.}
\label{fg.intro_layered}
\end{figure}

Consider a typical road scene as shown in Figure~\ref{fg.intro_layered} while driving a car. We first observe the immediate road, nearby obstacles (pedestrians, cars, cyclists, etc), followed by adjacent buildings and sky. These scene entities, or objects, can be layered in a typical road scene based on their locations. Understanding such a scene would require us to know the type of objects and spatial locations in the 3D world. Most conventional approaches look at this as two different problems: 3D reconstruction and object class segmentation. Recently, both these problems have been merged and solved as a single optimization problem. Along this avenue, several challenges exist. Prior segmentation algorithms focus on classifying each pixel individually into different semantic object classes. Such approaches are computationally expensive and may not respect the layered constraint that is preserved in most road scenes. In this paper, we jointly infer the semantic labels and depths of road scenes using the layered structure of street scenes.

In Figure~\ref{fg.intro_layered}, we use four layers to represent a street view image. The layers are ordered from the bottom of the image and they are associated with semantic classes. The first layer consists of only the ground. The second layer can have dynamic objects--vehicles, pedestrians, and cyclists. The third layer can only have buildings. The fourth layer can only contain sky. Each of these layers are supposed to model planar objects standing upright with respect to the ground at various distances from the camera. The transition between the layers happen at places where there is depth variation. In most road scenes, four layers are sufficient to model important object classes along with their layered depths. Our approach can fail in challenging scenarios with bridges or tunnels. However, these cases can be discovered by survey vehicles in an offline process. Autonomous vehicles can be alerted as we encounter these regions using GPS and we can use additional layers to correctly interpret such challenging regions.

We focus on obtaining layer-aware semantic labels and depths jointly from street-view images. We use a stereo camera setup and compute the disparity cost volume for depth cues. For obtaining semantic cues, we use a deep neural network, which extracts deep features from intensity images. The depth and semantic cues are formulated in an energy function that respects the layered street scene constraint. We propose an inference algorithm based on dynamic programming to efficiently minimize this energy function. Our inference algorithm is massively parallelizable. We develop a parallel implementation and achieve a 8.8 fps processing speed on GPU. Our method outperforms the competing algorithms that do not enforce the layered constraint. Our inference algorithm is general and can work on the data from other modalities including LIDAR and Radar sensors.

\subsection{Related work}

Our work is related to the general area of semantic segmentation and scene understanding, such as~\cite{Carreira2012semantic,Fulkerson2009,PhilippFFCRF,Shotton2009,Ess2009,Geiger2013,Wojek2013,scharwachter2014stixmantics}. While earlier approaches were based on hand-designed features, it has been shown recently that using deep neural networks for feature learning leads to better performance on this task~\cite{farabet2013learning,girshick2013rich,pinheiro2014,sharma2014recursive}.

The problem of jointly solving both semantic segmentation and depth estimation from stereo camera was addressed in~\cite{ladicky2010joint} as a unified energy minimization framework. Our work focuses on semantic labeling using ordering constraint on road scenes and using fewer classes applicable to road scenes. In~\cite{hoiem2005}, a typical road scene is classified into ground, vertical objects and sky to estimate the geometric layout from a single image. Objects like pedestrians and cars are segmented as vertical objects. This would be an under-representation for road scene understanding. \cite{felzenszwalb2010tiered} modeled the scene using two horizontal curves that divide the image into three regions: top, middle, and bottom.

One popular model for road scene is the stixel world that simplifies the world using a ground plane and a set of vertical sticks on the ground representing obstacles~\cite{badino2009}. Stixels are compact and efficient representation for upright objects on the ground. The stixel representation can simply be seen as the computation of two curves. The first curve runs on the ground plane enclosing the free space that can be immediately reached without collision, and the second curve encodes the vertical objects boundary. In order to compute the stixel world, either depth map from semi-global stereo matching algorithm (SGM)~\cite{hirschmuller2008} or cost volume~\cite{benenson2011} can be used. As with SGM, dynamic programming (DP) enables fast implementation for the computation of the stixels. Recently, ~\cite{yao2015} demonstrated a monocular free-space estimation using appearance cues.

Stix-mantics~\cite{scharwachter2014stixmantics}, a recently introduced model, gives more flexibility compared to stixels. Instead of having only one stixel for every column, they allow multiple segments along every column in the image and also combine nearby segments to form superpixel-style entities with better geometric meaning. Using these stixel-inspired superpixels, semantic class labeling is addressed.

We focus on obtaining layer-aware semantic labels and depths jointly from street-view images. Our work is closely related to many existing algorithms in vision, and most notably with tiered scene labeling~\cite{felzenszwalb2010tiered}, joint semantic segmentation and depth estimation~\cite{ladicky2010joint}, stixels and more recently, stix-mantics~\cite{scharwachter2014stixmantics}. Our approach achieves real-time processing speed and outperforms the competing algorithms~\cite{ladicky2010joint} in accuracy. We also achieve this performance without using explicit depth estimation and temporal constraints, which can be obtained using visual odometry. Similar to layered street view constraint, Manhattan constraints have been useful in indoor scene understanding~\cite{lee2009geometric,flint2011manhattan}.

The paper is organized as follow. In Section~\ref{sec::problem_formulation}, we present the problem formulation. Section~\ref{sec::features} discusses the extraction of depth and appearance features. The inference algorithm and its implementation are described in Section~\ref{sec::algorithm} and~\ref{sec::implementation}. Experiments are presented in Section~\ref{sec::experiments}.

\section{Layered Street View}\label{sec::problem_formulation}

\begin{figure}[!t]
\centering
\includegraphics[trim = 0.0in 1.9in 14.0in 0in, clip, width=3.1in]{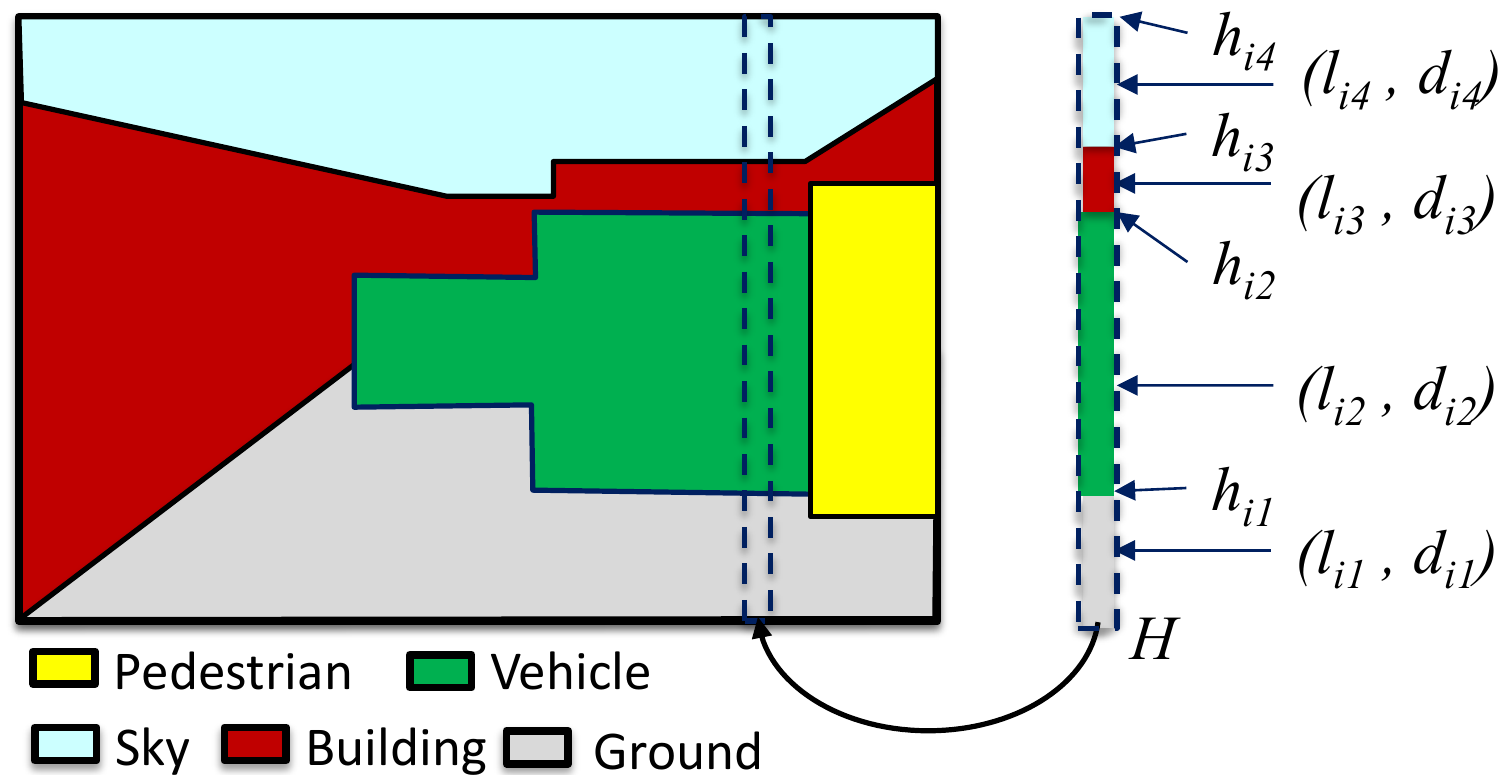}\\
\caption{{\bf Illustration of the layered street view problem formulation:} Each column of the image is divided up to four horizontal layers. The four layers are ordered from the bottom of the image. The model is compact and effective in representing a wide variety of scenarios in a typical road scene.}
\label{fig::problem_formulation}
\end{figure}

Our goal is to jointly estimate semantic labels and depth for each pixel in the street view image using both appearance and depth information. We adopt a layered image interpretation. An image is horizontally divided into four layers of different semantic and depth regions. The layers are ordered from the bottom to the top. In each image column, pixels belong to the same layer have the same semantic label and depth. The only exception is that depth of the pixels in the ground layer vary according to their vertical image coordinate, which is determined by the ground plane. The ground plane can either be obtained in an offline external calibration process or in an online estimation process such as using the v-disparity map~\cite{labayrade2002real}. We enforce a depth-order constraint, i.e. , depth of a lower layer is always smaller than depth of a higher layer in each image column.

\begin{figure*}[!t]
\centering
\includegraphics[trim = 0.0in 0in 0.0in 0in, clip, width=7.0in]{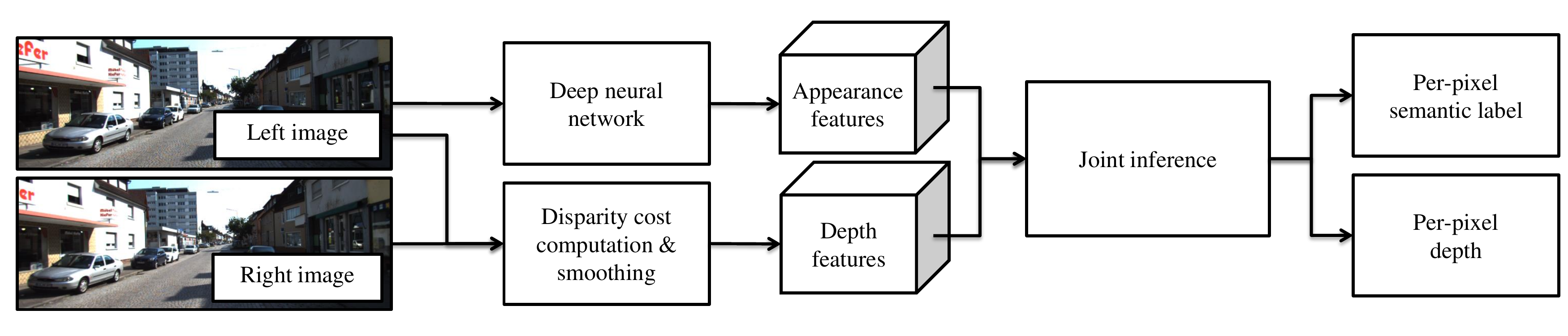}\\
\caption{{\bf Layered street view algorithm:} The proposed algorithm utilizes appearance features from a deep neural network and depth features from disparity costs. The features are used to jointly infer the dense semantic and depth labels.}
\label{fig::system_diagram}
\end{figure*}

In our four layer model, the first layer can only have ground. The second layer can have pedestrians or vehicles. The third layer can have only buildings. The fourth layer can only have sky. Note that we do not enforce that each image column has exactly four layers. A column can have any number of layers between one to four. If a layer is absent at a particular image column, the bottom of its upper layer and the top of its bottom layer are next to each other. This implies that the curves defining these layers need not be a smooth continuous one. The four-layer model provides a flexible method for enforcing geometry and semantics to the scene. The only assumption required is the planar world assumption, which is not restrictive for many applications requiring street scene understanding. If necessary, the geometry of the layered street scene model can be further enhanced to dense depth map with additional computational cost. Similarly, the model can be improved with more layers and semantic classes.

\subsection{Problem Formulation}

{\bf Notations:} We use $\mathcal{W}=\{ 1, 2,..., W \}$ and $\mathcal{H}=\{ 1, 2, ..., H\}$ to refer the sets that hold the horizontal $x$ and vertical $y$ coordinates respectively. We consider five different semantic object classes; namely, ground, vehicle, pedestrian, building, and sky. They are denoted by the symbols $\mathbb{G}$, $\mathbb{V}$, $\mathbb{P}$, $\mathbb{B}$, and $\mathbb{S}$ respectively. The set of the semantic class labels is denoted by $\mathcal{L} = \{ \mathbb{G}, \mathbb{V}, \mathbb{P}, \mathbb{B}, \mathbb{S}\}$. We use $\mathcal{D}$ for the set of disparity values. The words disparity and depth are used interchangeably in the paper for ease of presentation. It is understood that a one-to-one conversion can be easily obtained by using the parameters in the camera calibration matrix. The cardinality of the semantic label space and disparity values are denoted by $L=|\mathcal{L}|$ and $D=|\mathcal{D}|$ respectively.

We formulate the layered street view problem as a constrained energy minimization problem. The constraints encode the order of the semantic object class labels and depth values in each column. It limits the solution space of the variables associated with each image column. We solve the constrained energy minimization problem efficiently using an inference algorithm based on dynamic programming.

We use the variables, $h_{i1}$, $h_{i2}$, $h_{i3}$, and $h_{i4}$, to denote the $y$ coordinates of the top pixels of the four layers in the image column $i$. Let $l_{i1}$, $l_{i2}$, $l_{i3}$, and $l_{i4}$ be the semantic object class labels for the four layers and let $d_{i1}$, $d_{i2}$, $d_{i3}$, and $d_{i4}$ be the depths of the four layers in the image column $i$. The ordering constraint and the knowledge of the ground plane allow us to fix some parameters. The actual number of unknowns is only 5 given by $\mathbf{x}_i=[ h_{i1}, h_{i2}, h_{i3}, l_{i2}, d_{i3} ]$. Hence the label assignment for the entire image is given by $\mathbf{x}_{I} = [ \mathbf{x}_1, \mathbf{x}_2, ..., \mathbf{x}_W ]$. The number of possible assignments for an image column is in the order $O(H^3 L D)$ since $h_{i1},h_{i2},h_{i3} \in \mathcal{H}$, $l_{i2} \in \mathcal{L}$, and $d_{i3} \in \mathcal{D}$.

To rank the likelihood of the label assignment, we use evidence from image appearance features and stereo disparity matching features. We aggregate evidence from all the pixels in a column to compute the evidence. Let $U_i^A(\mathbf{x}_i)$ and $U_i^D(\mathbf{x}_i)$ be the data terms representing the semantic and depth label cost, respectively, incurred as assigning $\mathbf{x}_i$ to the image column $i$. The two terms are summed to yield the data term
\begin{align}
U_i(\mathbf{x}_i)=U_i^A(\mathbf{x}_i)+U_i^D(\mathbf{x}_i),
\end{align}
denoting the cost for assigning $\mathbf{x}_i$ to the image column $i$. Instead of working on the standard 2D Markov Random Field space where each pixel can have a depth value and a semantic label as independent variables, we reduce the problem to a constrained energy minimization problem given by
\begin{align}
    &\min_{ \mathbf{x}} \sum_{i=1}^{W} U_i(\mathbf{x}_i)\label{eqn::energy}\\
    \text{s.t. } &h_{i1} \ge h_{i2} \ge h_{i3} \ge h_{i4} = 1,\label{eqn::layer} \\
                 &d_{i1} < d_{i2} < d_{i3} < d_{i4},\label{eqn::depth_order}\\
                 &l_{i1} = \mathbb{G}, l_{i2} \in \{ \mathbb{V}, \mathbb{P}\}, l_{i3} = \mathbb{B}, l_{i4} = \mathbb{S}, \forall i\label{eqn::semantic_class}
\end{align}
where the constraint~(\ref{eqn::layer}) gives the layer structure, the constraint~(\ref{eqn::depth_order}) enforces the depth order, and the constraint~(\ref{eqn::semantic_class}) takes into account the possible semantic labels for each layer. The variable $d_{i2}$ is a function of $h_{i1}$, the top pixel location of the ground layer, because we assume the dynamic object is standing upright on the ground surface at the $h_{i1}$th row of the image. The energy function in Equation~(\ref{eqn::energy}) models the relation of pixels in the same column but not pixels in the same row. We use image patches centering around a pixel as the reception fields for the feature computation at the pixel location. Neighboring pixels have similar reception fields and thus have similar features.

The data term $U_i(\mathbf{x}_i)$ is the cost of assigning label $\mathbf{x}_i$ to the column $i$. It is the sum of the pixel-wise data terms given by
\begin{align}
U_i(\mathbf{x}_i) =
&\sum_{y=h_{i4}}^{h_{i3}-1} \big{(}E^A (i,y,\mathbb{S}) + E^D (i,y,\infty)\big{)}\nonumber\\
+&\sum_{y=h_{i3}}^{h_{i2}-1}\big{(}E^A (i,y,\mathbb{B}) + E^D (i,y,d_{i3})\big{)}\nonumber\\
+&\sum_{y=h_{i2}}^{h_{i1}-1}\big{(}E^A (i,y,l_{i2})     + E^D (i,y,d_{i2}(h_{i1}))\big{)}\nonumber\\
+&\sum_{y=h_{i1}}^{H}       \big{(}E^A (i,y,\mathbb{G}) + E^D (i,y,d_{i1})\big{)}
\end{align}
where the per pixel appearance data term $E^A(x,y,l)$ is the cost of assigning label $l$ to the pixel $(x,y)$ and the per pixel depth data term $E^D(x,y,d)$ is the cost of assigning depth $d$ to the pixel $(x,y)$. We use a deep neural network for obtaining the per pixel appearance data term $E^A(x,y,l)$ and use a standard disparity cost for obtaining the per pixel depth data term $E^D(x,y,d)$ detailed in Section~\ref{sec::features}. We summarize our layered street view algorithm in Figure~\ref{fig::system_diagram}.

\begin{figure*}[!t]
\centering
\includegraphics[trim = 0in 0.5in 0in 0in, clip, width=7.2in]{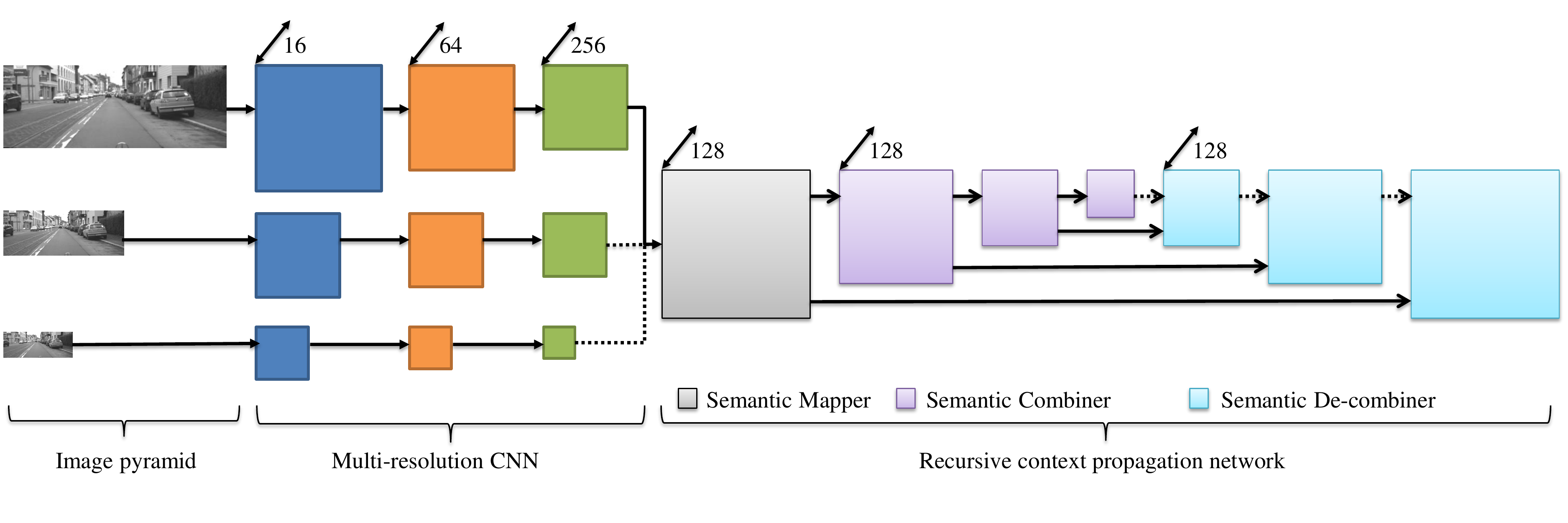}
\caption{{\bf Deep neural network for extracting the appearance features:} Our deep neural network contains two parts: multi-scale convolutional neural network and recursive context propagation network. The multi-scale convolutional neural network contains three convolutional layers and is applied to the Gaussian pyramid of the input image. It extracts multi-scale appearance features, which are fed into the recursive context propagation network. The recursive context propagation network consists of three sub-networks: the semantic mapper, the semantic combiner, and the semantic de-combiner networks. It embeds rich context information to the features and enhances their discriminative power. Note that the layers are color-coded. The same color is assigned to the layers sharing the filter weights. A dotted line indicates that the feature map is upsampled before feeding to the succeeding layer. For further details, please refer to Section~\ref{sec::features}.}
\label{fig::network}
\end{figure*}

\section{Features}\label{sec::features}
We rely on two types of features: depth and appearance.

\subsection{Depth}\label{sec::depth}
We use the smoothed absolute intensity difference for the per pixel depth data term, which is commonly used in stereo reconstruction algorithms. We first compute the pixel wise absolute intensity difference for each disparity value in $D$, which renders a cost volume representation. A box filter is then applied to smooth the cost volume. The per pixel depth data term is given by
\begin{align}
    E^D(x,y,d) = \frac{1}{N} \sum_{(\tilde{x},\tilde{y})\in P_{(x,y)}} | I_{L}(\tilde{x},\tilde{y} -d) - I_{R}(\tilde{x},\tilde{y})|
\end{align}
where $I_{L}$ and $I_{R}$ refer to the intensity values of the left and right images, $P_{(x,y)}$ is an image patch centered at $(x,y)$, and $N=|P_{(x,y)}|$ denotes the cardinality of the patch, serving as a normalization constant. The patch size is fixed to 11-by-11 in our experiments.

\subsection{Appearance}\label{sec::appearance}
We compute per pixel appearance data term using a deep neural netwrok. Our network, shown in Figure~\ref{fig::network}, consists of two parts: a multi-scale convolutional neural network (MSCNN) and a recursive context propagation network (RCPN). The MSCNN allows us to extract multi-scale appearance cues and RCPN allows us to extract rich contextual information.

{\bf MSCNN:}
Our MSCNN is a close variant to the neural network proposed in~\cite{farabet2013learning}. It has 3 convolutional layers. The first convolutional layer has 16 filters of size $8 \times 8 $ followed by rectified linear unit (ReLU) and $2 \times 2$ max-pooling processing. The second layer has 64 filters of size $7 \times 7$ followed by ReLU and $2 \times 2$ max-pooling processing. The third layer has 256 filters of size $7 \times 7$ followed by ReLU. The stacking of the convolutional layers yields a reception field of 47-by-47 pixels. Due to max-pooling, the resolution of the output feature map is smaller than that of the input image. We upsample the feature map to have the same resolution.

The 3-layer convolutional neural network (CNN) is applied separately to three scales of the Gaussian pyramid of the input image. Specifically, we downsample the input image using three different scales ($1$, $2$, and $4$) and use the 3-layer CNN to extract features at each scale. We use upsampling to bring all the feature maps to the same resolution. The feature maps from the three scales are concatenated to obtain the MSCNN feature map. Note that the filter weights are constrained to be the same for all the scales. The MSCNN extracts multi-scale appearance information for each pixel, which is then passed to the RCPN. For further details about MSCNN, please refer to~\cite{farabet2013learning}.

{\bf RCPN:}
Our RCPN is a variant of the network proposed in~\cite{sharma2014recursive}. It consists of three sub-networks: the semantic mapper network, the semantic combiner network, and the semantic de-combiner network. We use the RCPN to embed rich context information to the output appearance features. The semantic mapper network is a 1-layer CNN with 128 filters of size $1\times 1$ followed by ReLU. It maps each 768-dimensional feature of the MSCNN feature map to a 128-dimensional semantic feature, which is then fed into the semantic combiner network.

The semantic combiner network has three recursive layers, and each contains 128 filter of size $4 \times 4$ followed by ReLU. The semantic combiner fuses input features in a 4-by-4 region of the input semantic feature map to an output semantic feature. This process is non-overlapping; hence, each semantic combiner layer generates an output feature map that is 16 times smaller than the input one. Applying the 3-layer recursive combiner network renders an output feature map that is 4096 times smaller than the input feature map of the semantic mapper. The semantic combiner network recursively embeds context information from image regions with larger and larger spatial support. The output feature maps from the three layers form a context feature pyramid, which is fed into the semantic de-combiner network.

Similar to the semantic combiner network, the semantic de-combiner network has three recursive layers and each contains 128 filter of size $1 \times 1$ followed by ReLU. It is used to recursively distribute context information residing in the context pyramid back to the individual pixels, from higher to lower levels of the context pyramid. Each de-combiner layer fuses the feature map from the previous de-combiner layer with that from the corresponding level of the context pyramid. Note that our RCPN implementation differs from~\cite{sharma2014recursive} in two major places: 1) we use square patches for context propagation while~\cite{sharma2014recursive} uses superpixels\cite{liu2011entropy}, and 2) we use pyramids to represent hierarchy of context information while~\cite{sharma2014recursive} uses superpixel trees. Our design choices allow a more efficient implementation because we do not use superpixel segmentation.

{\bf Training:}
We use grayscale images. The pixel intensity values are scaled between 0 to 1 and centered by subtracting 0.5 before being fed into the deep neural network. To train the network, we connect the output layer to a fully connected layer having 5 neurons, corresponding to ground, pedestrian, vehicle, building, and sky classes. The fully connected layer is followed by the softmax layer. The network is trained by minimizing the cross-entropy error via stochastic gradient descent with momentum. We use the Caffe library~\cite{jia2014caffe} for training. The number of pixels in the semantic classes can be quite different. To avoid the bias from dominant classes (ground and building), we weight the cross-entropy loss based on the semantic class distribution, which yields better performance in practice.

We use the negative logarithm of the softmax scores of the semantic classes as the per-pixel appearance data terms. Let $f(x,y,l)$ be the softmax score of the deep neural network at pixel location $(x,y)$, which represents the probability of assigning the label $l$ to the pixel. The per pixel appearance data term is given by
\begin{align}
    E^A(x,y,l) = -\beta\log f(x,y,l)
\end{align}
where $\beta$ is a parameter controlling the relative weight of the appearance and depth data terms.

\section{Efficient Inference Algorithm}\label{sec::algorithm}

We decompose the energy minimization problem in Equation~(\ref{eqn::energy}) into $W$ sub-problems where the $i$th sub-problem is given by
\begin{align}
    &\min_{ \mathbf{x}_i} U_i(\mathbf{x}_i)\label{eqn::subproblem}\\
    \text{s.t. } &h_{i1} \ge h_{i2} \ge h_{i3} \ge h_{i4} = 1\\
                 &d_{i1} < d_{i2} < d_{i3} < d_{i4}\\
                 &l_{i1} = \mathbb{G}, l_{i2} \in \{ \mathbb{V}, \mathbb{P}\}, l_{i3} = \mathbb{B}, l_{i4} = \mathbb{S}.
\end{align}
We solve each of the sub-problems optimally and combine their solutions to construct the semantic labeling and depth map of the image. For simplicity, we will drop the subscript $i$ in the discussion below.

Each of the sub-problems can be mapped to a 1D chain labeling problem. The chain has $4$ nodes where the first node contains the variables $( h_{1}, d_{1}, l_{1} )$, the second node contains the variables $( h_{2}, d_{2}, l_{2} )$, the third node contains the variables $( h_{3}, d_{3}, l_{3} )$, and the fourth node contain the variables $( h_{4}, d_{4}, l_{4} )$. Utilizing the recursion in the label cost evaluation, a standard dynamic programming algorithm can solve the inference on the 1D chain with a complexity of $O( (HDL\cdot H)^2 )$ where the product $HDL$ represents the size of label space at each node and the second $H$ comes from the label cost evaluation at each node. Unfortunately, the complexity is too high for real-time applications.

We propose a variant of the dynamic programming algorithm to reduce the complexity of solving the sub-problem in~(\ref{eqn::subproblem}) to $O(H^2D)$ and achieve real-time performance. We first note that some of the variables are known from our street view setup as discussed in the problem formulation section. We only need to search the values for $h_{1}, h_{2}, h_{3}, l_{2}, d_{3}$. For any combination of $h_1$ $h_2$ and $l_2$, we need to find the best combination of $d_3$ and $h_3$. In the following, we show that pre-computing the best combination of $d_3$ and $h_3$ for any $h_1$ $h_2$ and $l_2$ can be achieved in $O(H^2D)$ time using recursion.

We first observe that the problem in~(\ref{eqn::subproblem}) can then be written as
\begin{align}
\min_{h_1,h_2,l_2}
&\sum_{y=h_{2}}^{h_{1}-1} \big{(}E^A (i,y,l_{2})     + E^D (i,y,d_{2}(h_{1}))\big{)} +
\nonumber\\
&\sum_{y=h_{1}}^{H}       \big{(}E^A (i,y,\mathbb{G}) + E^D (i,y,d_{1}) + Q_i(h_1,h_2)\big{)}
\end{align}
where $Q_i$ is an intermediate cost table given by
\begin{align}
 &Q_i(h_1,h_2) \equiv \min_{\substack{h_3,d_3\\d_3>d_2(h_1)}}
\sum_{y=h_{4}}^{h_{3}-1} \big{(}E^A (i,y,\mathbb{S}) \nonumber\\
&+ E^D (i,y,\infty)\big{)} + \sum_{y=h_{3}}^{h_{2}-1} \big{(}E^A (i,y,\mathbb{B}) + E^D (i,y,d_{3})\big{)}
\end{align}
Note that depth of the second layer object $d_2$ is a function of $h_1$ because $d_2$ can be uniquely determined from $h_1$ and the ground plane equation. As a result, $Q_i$ depends on both $h_1$ and $h_2$.

By integrating $E^A (i,y,\mathbb{B}) + E^D (i,y,d_3)$ and $E^A (i,y,\mathbb{S}) + E^D (i,y,\infty)$ along the $y$ direction, the sum given by
\begin{align}
R_i(h_2,h_3,d_3) \equiv &\sum_{y=h_{4}}^{h_{3}-1} \big{(}E^A (i,y,\mathbb{S}) + E^D (i,y,\infty)\big{)}+\nonumber\\
&\sum_{y=h_{3}}^{h_{2}-1} \big{(}E^A (i,y,\mathbb{B}) + E^D (i,y,d_{3})\big{)}
\end{align}
for all combination of $h_2$ and $h_3$ can be computed in $O(H^2)$ time for each $d_3$. We further note that $Q_i$ can be computed via a recursive update rule given by
\begin{align}
&Q_i(h_1+1,h_2) = \min( Q_i(h_1,h_2),\space\min_{h_3} R_i(h_2,h_3,\tilde{d}) )
\end{align}
where $\tilde{d}$ is an integer satisfying $d_2 (h_1) \leq \tilde{d} < d_2 (h_1+1)$ and is used to ensure that the depth ordering constraint between the second and third layers is met. Intuitively, we are computing a running min structure along the decreasing depth of the building layer. The recursive update rule allows us to compute $Q_i$ for any $h_1$ and $h_2$ in $O(H^2D)$ time. As a result, the complexity of finding the best configuration for a partition can be reduced to $O(H^2L + H^2D) = O(H^2D)$ where $H^2L$ is the time required for searching combinations of $h_1$ $h_2$ and $l_2$. We perform the 1D labeling algorithm to each image column. The overall complexity of the labeling algorithm is $O(WH^2D)$.

\section{Implementation}
\label{sec::implementation}

Our algorithm is massively parrallelizable and can be implemented using CUDA, a general purpose parallel computing language for NVIDIA GPUs~\cite{cuda2014programming}. A GPU comprises a large number of Single-Instruction-Multiple-Data (SIMD) processor cores to allow many threads to execute common operations concurrently on large data arrays. In our implementation of the labeling algorithm, we exploit data level parallelism in all stages of computation.

\begin{itemize}
\item[1] {\bf Depth data term:} We use $W \times H$ threads to compute the disparity values for each pixel at $(x,y)$. We implemented the box filter using the sliding window approach. First, we use $H$ threads to perform 1D sliding window on each row. As the window moves from left to right in the horizontal direction, the new pixel value on the right is added and the existing one on the left is subtracted. We then use $W$ threads to carry out the same computation for each column.
\item[2] {\bf Appearance data term:} We use the Caffe library~\cite{jia2014caffe} to compute the softmax scores output of the deep neural network. We use NVIDIA's cuDNN library to achieve additional speed up.
\item[3] {\bf Intermediate cost table:} We first compute the integral of $E^A (i,y,\mathbb{B}) + E^D (i,y,d_{3})$ and $E^A (i,y,\mathbb{S}) + E^D (i,y,\infty)$ over $y$ such that the two sum terms can be retrieved in constant time for any range. This can be done in parallel for each column $i$. We observe that for a fixed $h_2$, computing $R_i(h_2,h_3,d_3)$ over each $i$ and $h_3$ can be jointly parallelized. Therefore we use $h_2 \times W$ threads to compute an intermediate table for $d_3$ and $h_3$ and use $W$ threads to find the combinations that yield the minimum cost for each $h_1$ and $h_2$.
\item[4] {\bf Energy minimization and labeling:} Using previously computed $Q$, we use $W$ threads to search the $\mathbf{x}_i$ with the minimum cost in each column in parallel.
\end{itemize}

Memory layout is an important factor for processing speed in GPU. By default, our image data is stored in row-major form. The GPU implementation naturally takes advantage of this memory layout by assigning threads to work on pixels on the same row, which resides in memory as a continuous array. This allows GPU to coalesce the memory accesses of the threads such that the GPU memory bandwidth is efficiently utilized. In addition, our algorithm avoids reshaping or transposing the data in the memory, which would take extra memory and time.

We execute our algorithm on a Windows 7 desktop computer equipped with NVIDIA Tesla K40 GPU along with Intel i7 processor. We set the size of one-dimension thread blocks to be 64 and the size of two-dimension thread blocks to be $32 \times 32$ to facilitate efficient scheduling of the threads on target GPU. To avoid register spills to local memory, we minimize local variable declarations. In our algorithm, no data needs to be shared among threads within a block, therefore shared memory is not used in the implementation.

\begin{table*}
\caption{The table compares the proposed method with several competing algorithms on semantic object class labeling accuracy using the Daimler Urban Segmentation Dataset. The results are shown in percentage using the IoU measure.}
\centering{\begin{tabular}{|c|cccc|cc|}
\hline
{Class}$\backslash${Method}&
\begin{tabular}{@{}c@{}}Joint-Opt.\\ALE~\cite{ladicky2010joint}\end{tabular} &
\begin{tabular}{@{}c@{}}Stix-mantics\\~\cite{scharwachter2014stixmantics}\end{tabular}&
\begin{tabular}{@{}c@{}}Darwin-pairwise\\~\cite{gould2012darwin}\end{tabular} &
\begin{tabular}{@{}c@{}}PN-RCPN\\~\cite{sharma2015deep}\end{tabular} &
\begin{tabular}{@{}c@{}}Appearance\\ features \end{tabular}&
Proposed \\
\hline
Ground         & 94.9 & 93.8 & 95.7 & {\bf96.7} & {\bf96.7} & 96.4 \\
Vehicle        & 76.0 & 78.8 & 68.7 & 79.4 & 80.7 & {\bf83.3} \\
Pedestrian     & {\bf73.1} & 66.0 & 21.2 & 68.4 & 61.3 & 71.1 \\
Sky            & {\bf95.5} & 75.4 & 94.2 & 91.4 & 87.6 & 89.5 \\
Building       & 90.6 & 89.2 & 87.6 & 86.3 & 87.4 & {\bf91.2} \\
\hline
Avg (all)      & 86.0 & 80.6 & 73.5 & 84.5 & 82.8 & {\bf86.3} \\
Avg (dynamic)  & 74.5 & 72.4 & 44.9 & 73.8 & 71.0 & {\bf77.2} \\
\hline
Runtime per frame in second & 111 & 0.05 & N/A & 2.8 & 0.07 & 0.11 \\
\hline
\end{tabular}}\label{tbl::semantic_comparison}
\end{table*}
\begin{figure*}[!t]
\centering
\includegraphics[trim = 0in 0in 0in 0in, clip, width=7.2in]{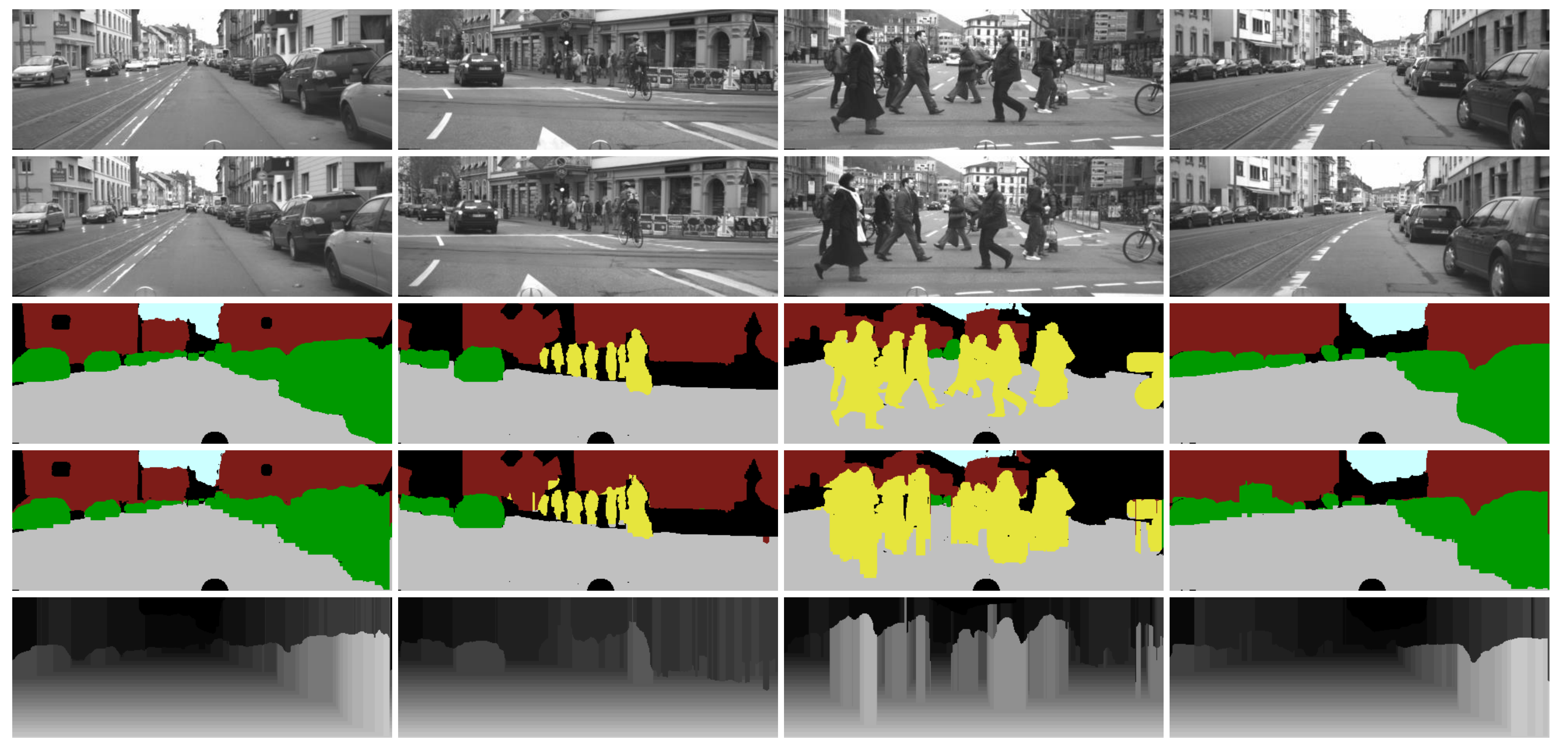}
\caption{{\bf Visualization:} The figure visualizes the output computed from the proposed method. From top to bottom, we show the left images, the right images, the ground truth semantic labeling, the semantic labeling, and depth. The black regions are the regions where the ground truths are not available.}
\label{fig::visualization}
\end{figure*}

\section{Experiments}\label{sec::experiments}

{\bf Benchmark:} We evaluated our approach using the public Daimler Urban Segmentation dataset \cite{scharwachter2013efficient}. The dataset contains 500 stereo grayscale image pairs with pixel-wise semantic class annotations for the left images. While the image size in the dataset is 1024x440, only the middle region from $(24,40)$ to $(1000,400)$ is fully labeled. Hence, the effective image size is 976x360. The dataset is composed for evaluating only the semantic labeling using stereo image pairs. There is no ground truth for depth.

The semantic labels in the annotations include ground, sky, building, pedestrian, vehicle, curbs, bicyclist, motorcyclist, and background clutters. However, only the ground, sky, building, pedestrian, and vehicle are considered in the evaluation protocol. The performance metric for the semantic labeling is based on the PASCAL VOC intersection over union (IoU) measure~\cite{everingham2010pascal}, which is the ratio of cardinality of the intersection of the ground truth and estimated semantic segments over that of their union. Let $\tilde{S}$ and $S$ be the set of pixels labeled as sky in the computed semantic class label map and the ground truth label map, respectively. The IoU measure of the sky class is given by
\begin{equation}
IoU(\text{Sky}) = \frac{|\tilde{S}\cap S|}{|\tilde{S}\cup S|}.
\end{equation}
The larger the IoU measure, the better the matching between the ground truth and estimated segments; and, hence, the better the semantic labeling accuracy.

{\bf Evaluation:} We followed the evaluation protocol described in \cite{scharwachter2014stixmantics}, which used the first 300 stereo image pairs in the dataset for training and the remaining 200 stereo image pairs for testing. During testing, we downsampled the input images by half in each dimension. This was necessary for our GPU implementation. During evaluation, we upsampled the image to the original size.

We use left images of the stereo pairs for training our deep neural network. During testing, the network outputs per-semantic-class softmax scores for each pixel. We compared our approach with several approaches. The competing approaches include the joint-optimal ALE (ALE) algorithm~\cite{ladicky2010joint}, the stix-mantics~\cite{scharwachter2014stixmantics}, the Darwin pairwise~\cite{gould2012darwin}, and the PN-RCPN~\cite{sharma2015deep}. The algorithms in~\cite{ladicky2010joint,sharma2015deep} utilizes superpixel segmentations as input, which demands additional computation resources. The stix-mantics algorithm~\cite{scharwachter2014stixmantics} uses depth, obtained using an FPGA chip, and temporal constraints from adjacent stereo images. We achieve better accuracy and comparable computational performance without using an FPGA chip and temporal constraints.

In Table~\ref{tbl::semantic_comparison}, we compare the semantic labeling accuracy of the competing algorithms. The results of the competing algorithms are duplicated from the Daimler dataset website~\cite{SceneLabeling6D}. Note that the results in the website are different from those reported in the original paper~\cite{scharwachter2014stixmantics} because the unlabeled pixels were initially not excluded from the IoU computation in the original paper~\cite{scharwachter2014stixmantics}.

{\bf Performance:} In the table, we show that our method achieves the state-of-the-art performance of 86.3\%, while ALE achieves an accuracy of 86.0\%. The stix-mantics algorithm achieves an accuracy of 80.6\%, which is significantly lower than our method and ALE. In terms of the performance for the dynamic objects (vehicles and pedestrians), the proposed algorithm achieves an accuracy of 77.2\%, outperforming ALE, which gets 74.5\%. In terms of speed, we are several magnitude faster than ALE. We generate both depth and semantic labels in 114.1 ms, while ALE takes 111,000 ms. The stix-mantics algorithm is slightly faster than our method requiring only 50 ms. However, they use an FPGA chip to precompute the depth before estimating the semantic labels, whereas we jointly compute both semantic labels and depth.

In Figure~\ref{fig::visualization}, we visualize the semantic labels and depth from the proposed algorithm. Qualitatively, we find that we obtain visually accurate semantic labels and our depth map resembles a piece-wise planar approximation of the 3D scene.

\begin{figure}[!t]
\centering
\includegraphics[trim = 0in 0in 0in 0in, clip, width=3.5in]{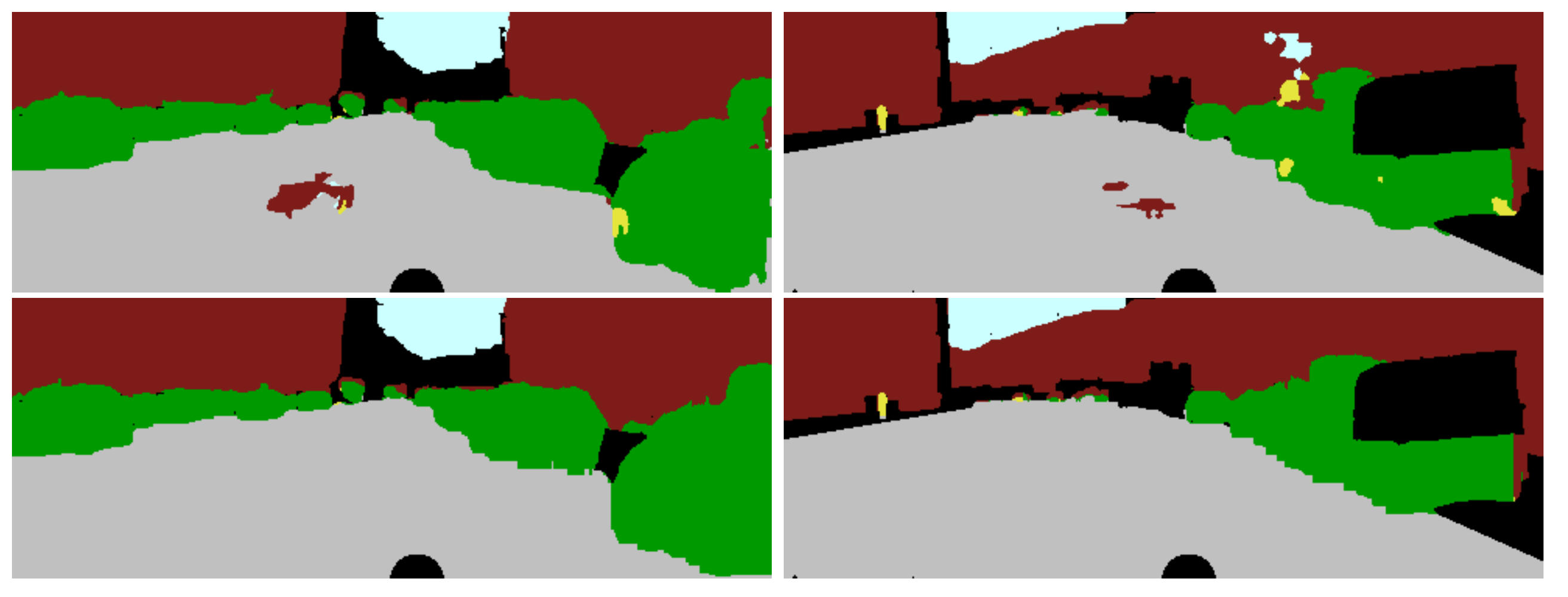}\\
\caption{{\bf The advantages of the layered constraint:} Top: Semantic labeling using only the appearance features. Bottom: Semantic labeling using the layered constraint. This constraint avoids geometrically impossible labeling, which may occur while using only appearance features extracted using the deep neural network.}
\label{fig::compare_deep}
\end{figure}

We observe that our appearance features alone performs quite well. It achieves an average accuracy of 82.8\%. The layered constraint allows us to avoid impossible labelings such as having ground regions in the middle of sky, or having vehicles in the middle of a building, etc. By incorporating this constraint, our performance improves to 86.3\%, which corresponds to a 20.3\% error reduction. This improvement can be qualitatively seen in a few examples in Figure~\ref{fig::compare_deep}.

\begin{table}
\caption{Algorithm Component execution time}
\centering{
\begin{tabular}{|c|c|}
\hline
Component name         & Execution time in ms \\
\hline
Depth data term (cost volume)   & 5.2 \\
Appearance data term (DNN)   	& 70.0 \\
Intermediate table $Q$ 			& 25.6 \\
Inference 			   			& 13.3 \\
\hline
Overall							& 114.1 \\
\hline
\end{tabular}}\label{tbl::time}
\end{table}

We report the execution time required by the individual steps in our algorithm in Table~\ref{tbl::time}. All the computation is performed in the GPU. Overall, our algorithm takes 114.1 ms to infer the semantic labels and depth. The run time can be further reduced by half, by utilizing a second GPU card for processing the neural network computation.

\section{Conclusion}\label{sec::concl}

We propose a novel layered street view model and develop an efficient algorithm to jointly estimate semantic labels and depth for street view images. We obtain this result using appearance features, which can be computed from a deep neural network, and depth features, which can be derived from stereo disparity costs. Our algorithm outperforms the competing methods on the Daimler Urban Segmentation data set and runs at 8.8 frame-per-second using a GPU implementation.

\section*{Acknowledgement}
We thank Jay Thornton, Shumpei Kameyama, reviewers, and area chairs for their feedback and support of this work.

\bibliographystyle{ieee}
\bibliography{layered_arxiv}

\begin{thebibliography}{10}\itemsep=-1pt

\bibitem{SceneLabeling6D}
Daimler urban segmentation dataset, 2015.

\bibitem{badino2009}
H.~Badino, U.~Franke, and D.~Pfeiffer.
\newblock The stixel world - a compact medium level representation of the
  3d-world.
\newblock In {\em DAGM}, 2009.

\bibitem{benenson2011}
R.~Benenson, R.~Timofte, and L.~Gool.
\newblock Stixels estimation without depth map computation.
\newblock In {\em IEEE International Conference on Computer Vision}, 2011.

\bibitem{Carreira2012semantic}
J.~Carreira, R.~Caseiro, J.~Batista, and C.~Sminchisescu.
\newblock Semantic segmentation with second-order pooling.
\newblock In {\em European Conference on Computer Vision}, 2012.

\bibitem{cuda2014programming}
N.~Corporation.
\newblock {NVIDIA} {CUDA} {C} programming guide.
\newblock 2014.

\bibitem{Ess2009}
A.~Ess, T.~Mueller, H.~Grabner, and L.~Gool.
\newblock Segmentation-based urban traffic scene understanding.
\newblock In {\em British Machine Vision Conference}, 2009.

\bibitem{everingham2010pascal}
M.~Everingham, L.~Van~Gool, C.~K. Williams, J.~Winn, and A.~Zisserman.
\newblock The pascal visual object classes (voc) challenge.
\newblock {\em International journal of computer vision}, 88(2):303--338, 2010.

\bibitem{farabet2013learning}
C.~Farabet, C.~Couprie, L.~Najman, and Y.~LeCun.
\newblock Learning hierarchical features for scene labeling.
\newblock {\em IEEE Transactions on Pattern Analysis and Machine Intelligence},
  35(8):1915--1929, 2013.

\bibitem{felzenszwalb2010tiered}
P.~F. Felzenszwalb and O.~Veksler.
\newblock Tiered scene labeling with dynamic programming.
\newblock In {\em IEEE Conference on Computer Vision and Pattern Recognition},
  pages 3097--3104, 2010.

\bibitem{flint2011manhattan}
A.~Flint, D.~Murray, and I.~Reid.
\newblock Manhattan scene understanding using monocular, stereo, and 3d
  features.
\newblock In {\em IEEE International Conference on Computer Vision}.

\bibitem{Fulkerson2009}
B.~Fulkerson, A.~Vedaldi, and S.~Soatto.
\newblock Class segmentation and object localization with superpixel
  neighborhoods.
\newblock In {\em IEEE International Conference on Computer Vision}, 2009.

\bibitem{Geiger2013}
A.~Geiger, M.~Lauer, C.~Wojek, C.~Stiller, and R.~Urtasun.
\newblock 3d traffic scene understanding from movable platforms.
\newblock In {\em IEEE Transactions on Pattern Analysis and Machine
  Intelligence}, 2013.

\bibitem{girshick2013rich}
R.~Girshick, J.~Donahue, T.~Darrell, and J.~Malik.
\newblock Rich feature hierarchies for accurate object detection and semantic
  segmentation.
\newblock {\em arXiv preprint arXiv:1311.2524}, 2013.

\bibitem{gould2012darwin}
S.~Gould.
\newblock Darwin: A framework for machine learning and computer vision research
  and development.
\newblock {\em The Journal of Machine Learning Research}, 13(1):3533--3537,
  2012.

\bibitem{hirschmuller2008}
H.~Hirschmuller.
\newblock Stereo processing by semiglobal matching and mutual information.
\newblock {\em IEEE Transactions on Pattern Analysis and Machine Intelligence},
  2008.

\bibitem{hoiem2005}
D.~Hoiem, A.~A. Efros, and M.~Hebert.
\newblock Automatic photo pop-up.
\newblock {\em ACM Transactions on Graphics}, 2005.

\bibitem{jia2014caffe}
Y.~Jia, E.~Shelhamer, J.~Donahue, S.~Karayev, J.~Long, R.~Girshick,
  S.~Guadarrama, and T.~Darrell.
\newblock Caffe: Convolutional architecture for fast feature embedding.
\newblock {\em arXiv preprint arXiv:1408.5093}, 2014.

\bibitem{PhilippFFCRF}
P.~Krahenbuhl and V.~Koltun.
\newblock Efficient inference in fully connected crfs with gaussian edge
  potentials.
\newblock In {\em Neural Information Processing Systems}, 2011.

\bibitem{labayrade2002real}
R.~Labayrade, D.~Aubert, and J.-P. Tarel.
\newblock Real time obstacle detection in stereovision on non flat road
  geometry through" v-disparity" representation.
\newblock In {\em Intelligent Vehicle Symposium}, volume~2, pages 646--651.
  IEEE, 2002.

\bibitem{ladicky2010joint}
L.~Ladicky, P.~Sturgess, C.~Russell, S.~Sengupta, Y.~Bastanlar, W.~Clocksin,
  P.~H. Torr, et~al.
\newblock Joint optimisation for object class segmentation and dense stereo
  reconstruction.
\newblock In {\em British Machine Vision Conference}, 2010.

\bibitem{lee2009geometric}
D.~C. Lee, M.~Hebert, and T.~Kanade.
\newblock Geometric reasoning for single image structure recovery.
\newblock In {\em IEEE Conference on Computer Vision and Pattern Recognition}.

\bibitem{liu2011entropy}
M.-Y. Liu, O.~Tuzel, S.~Ramalingam, and R.~Chellappa.
\newblock Entropy rate superpixel segmentation.
\newblock In {\em IEEE Conference on Computer Vision and Pattern Recognition},
  2011.

\bibitem{pinheiro2014}
P.~Pinheiro and R.~Collobert.
\newblock Rich feature hierarchies for accurate object detection and semantic
  segmentation.
\newblock {\em International Conference on Machine Learning}, 2014.

\bibitem{scharwachter2013efficient}
T.~Scharw{\"a}chter, M.~Enzweiler, U.~Franke, and S.~Roth.
\newblock Efficient multi-cue scene segmentation.
\newblock In {\em Pattern Recognition}, pages 435--445. Springer, 2013.

\bibitem{scharwachter2014stixmantics}
T.~Scharw{\"a}chter, M.~Enzweiler, U.~Franke, and S.~Roth.
\newblock Stixmantics: A medium-level model for real-time semantic scene
  understanding.
\newblock In {\em European Conference on Computer Vision}, pages 533--548.
  Springer, 2014.

\bibitem{sharma2015deep}
A.~Sharma, O.~Tuzel, and D.~W. Jacobs.
\newblock Deep hierarchical parsing for semantic segmentation.
\newblock In {\em IEEE Conference on Computer Vision and Pattern Recognition}.
  2015.

\bibitem{sharma2014recursive}
A.~Sharma, O.~Tuzel, and M.-Y. Liu.
\newblock Recursive context propagation network for semantic scene labeling.
\newblock In {\em Neural Information Processing Systems}, pages 2447--2455,
  2014.

\bibitem{Shotton2009}
J.~Shotton, J.~Winn, C.~Rother, and A.~Criminisi.
\newblock Textonboost for image understanding:multi-class object recognition
  and segmentation by jointly modeling texture, layout, and context.
\newblock In {\em International Journal of Computer Vision}, 2009.

\bibitem{Wojek2013}
C.~Wojek, S.~Walk, S.~Roth, K.~Schindler, and B.~Schiele.
\newblock Monocular visual scene understanding: Understanding multi-object
  traffic scenes.
\newblock {\em IEEE Transactions on Pattern Analysis and Machine Intelligence},
  2013.

\bibitem{yao2015}
J.~Yao, S.~Ramalingam, Y.~Taguchi, Y.~Miki, and R.~Urtasun.
\newblock Estimating drivable collision-free space from monocular video.
\newblock In {\em Winter Conference on Applications of Computer Vision}, 2015.

\end{thebibliography}
\end{document}